\documentclass{article}

\usepackage{aaai}
\usepackage{latexsym}
\usepackage{amstext,amssymb}

\newtheorem{theorem}{Theorem}

\newcommand{\nameset}{{\ensuremath{N}}}
\newcommand{\mameset}{{\ensuremath{M}}}
\renewcommand{\nameset}{{\cal N}}
\renewcommand{\mameset}{{\cal M}}

\newtheorem{definition}{Definition}
\newtheorem{corollary}{Corollary}

\newcommand{\oksym}{{\mathsf{ok}}}

\newcommand{\blockedsym}{{\mathsf{bl}}}
\newcommand{\appliedsym}{{\mathsf{ap}}}
\newcommand{\insym}{{\mathsf{in}}}

\newcommand{\ok}[1]{\ensuremath{\oksym(#1)}}
\newcommand{\oko}[2]{\ensuremath{\oksym'(#1,#2)}} 

\newcommand{\blocked}[1]{\ensuremath{\blockedsym(#1)}}

\newcommand{\applied}[1]{\ensuremath{\appliedsym(#1)}}

\newcommand{\inp}[2]{\ensuremath{\insym(#1,#2)}}

\newcommand{\LPif}{\leftarrow}

\newcommand{\To}[1]{\ensuremath{T_{#1}}}
\newcommand{\T}[2]{\To{#1}#2}
\newcommand{\Tindo}[2]{\To{#2}^{#1}}
\newcommand{\Tind}[3]{\Tindo{#1}{#2}#3}
\newcommand{\head}[1]{\mathit{head}(#1)}
\newcommand{\pbody}[1]{\mathit{body}^+(#1)}
\newcommand{\nbody}[1]{\mathit{body}^-(#1)}
\newcommand{\body}[1]{\mathit{body}(#1)}
\newcommand{\nafo}[0]{\mathit{not}}
\newcommand{\naf}[0]{\nafo\;}

\newcommand{\reduct}[2]{#1^{#2}}
\newcommand{\reductr}[1]{#1^{+}}

\newcommand{\Tho}[0]{\mbox{\rm Cn}}
\newcommand{\Th}[1]{\Tho(#1)}
\newcommand{\iec}[0]{i.e.,\ }
\newcommand{\commadots}[0]{,\ldots ,}
\newcommand{\PREC}[2]{#1\prec #2}
\newcommand{\PRECio}[1]{<_{#1}}
\newcommand{\PRECi}[3]{#2\PRECio{#1} #3}
\newcommand{\PRECMo}[0]{<}
\newcommand{\PRECM}[2]{\ensuremath{#1\PRECMo#2}}

\newcommand{\GR}[2]{\ensuremath{{\Gamma}_{#1}^{#2}}} 

\newcommand{\ap}[2]   {\ensuremath{{a_{#1}(#2   )}}}   
\newcommand{\bl}[3]   {\ensuremath{{b_{#1}(#2,#3)}}}   
\newcommand{\cokt}[2] {\ensuremath{{c_{#1}}(#2)}}       
\newcommand{\cok}[3]  {\ensuremath{{c_{#1}(#2,#3)}}}   

\newcommand{\Lit}[0] {\mathit Lit}

\newcommand{\nameo}[0]{\mathit{n}}
\newcommand{\namef}[1]{\nameo(#1)}
\newcommand{\name}[1]{\nameo_{#1}}

\newcommand{\egc}[0]{e.g.,\ }
\newcommand{\ie}[0]{i.e.\ }

\title{Logic Programs with Compiled Preferences\thanks{This work was 
partially supported by the Austrian Science Fund Project N Z29-INF.}}

\author{
   James P.\ Delgrande\\
     School of Computing Science, \\
     Simon Fraser University, \\
     Burnaby, B.C., \\
     Canada  V5A 1S6, \\
     jim@cs.sfu.ca,
   \And
   Torsten Schaub \\
     Institut f\"ur Informatik, \\
     Universit\"at Potsdam, \\
     Postfach 60 15 53, \\
     D--14415 Potsdam,
     Germany, \\
     torsten@cs.uni-potsdam.de 
   \And
   Hans Tompits \\
     Abt.\ Wissensbasierte Systeme, \\
     Technische Universit\"at Wien, \\
     Favoritenstra{\ss}e~9--11, \\
     A--1040 Wien, Austria, \\
     tompits@kr.tuwien.ac.at
 }

\date{}
\begin{document}

\maketitle
\bibliographystyle{aaai}

\begin{abstract}
We describe an approach for compiling preferences into logic programs
under the answer set semantics.
An {\em ordered} logic program is an extended logic program in which
rules are named by unique terms, and in which preferences among rules
are given by a set of atoms of the form $s \prec t$ where $s$ and $t$
are names.
An ordered logic program is transformed into a second,
regular, extended logic program wherein the preferences are respected,
in that the answer sets obtained in the transformed theory correspond
with the preferred answer sets of the original theory.
Our approach allows both the specification of \emph{static} orderings (as
found in most previous work), in which preferences are external to a logic
program, as well as orderings on \emph{sets} of rules.
In large part then, we are interested in describing a general
{\em methodology} for uniformly incorporating preference information in a
logic program.
Since the result of our translation is an extended logic program, we can make
use of existing implementations, such as \texttt{dlv} and \texttt{smodels}.
To this end, we have developed a compiler, available on the web, as a
front-end for these programming systems.
\end{abstract}

\section{Introduction}\label{sec:introduction}

In commonsense reasoning one frequently prefers one outcome over
another, or the application of one rule over another, or the drawing of one
default conclusion over another.
For example, in buying a car one may have various desiderata in mind
(inexpensive, safe, fast, etc.) where these preferences come in varying
degrees of importance.
In legal reasoning, laws may apply by default but the laws themselves may
conflict.
So municipal laws will have a lower priority than state laws, and newer laws
will take priority over old.
Further, if these preferences conflict, there will need to be higher-order
preferences to decide the conflict.

In this paper we explore the problem of preference orderings within the
framework of 
extended logic programs under the answer set semantics~\cite{gellif91a}.
The general methodology was first proposed in \cite{delsch97a}, in
addressing preferences in default logic.
Previous work in dealing with preferences has for the most part treated
preference information at the \emph{meta-level\/}
(see next to the last section
for a discussion of previous approaches).
In contrast, we remain within the framework of extended logic programs:
We begin with an {\em ordered} logic program, which is an extended
logic program in which rules are named by unique terms and in which
preferences among rules are given by a new set of atoms of the form
$s \prec t$, where $s$ and $t$ are names.
Thus, preferences among rules are encoded at the {\em object-level}.
An ordered logic program is transformed into a second, regular,
extended logic program wherein the preferences are respected, in the sense
that the answer sets obtained in the transformed theory correspond to
the preferred answer sets of the original theory.
The approach is sufficiently general to allow the specification of preferences
among preferences, preferences holding in a particular context, and
preferences holding by default.

Our approach can be seen as a general {\em methodology} for
uniformly incorporating preference information within a logic program.
This transformational approach has several advantages.
First, it is flexible.
So one can encode how a preference order interacts with other information,
or how different types of preference orders (such as specificity,
authority, recency, etc.) are to be integrated.
Second, it is easier to compare differing approaches handling such
orderings, since they can be represented uniformly in the same general
setting.
Thus, for instance, if someone doesn't like the notion of preference
developed here, they may encode their own within this framework.
Lastly, it is straightforward implementing our approach:
In the present case, we have developed a translator for ordered logic
programs that serves as a front-end for the logic programming systems
\texttt{dlv}~\cite{dlv97a} and \texttt{smodels}~\cite{niesim97a}.

The next section gives background terminology and notation.
Afterwards, our central approach is described, followed by an exploration of
its formal properties. 
We then continue with an overview of further features and
extensions, and provide a pointer to the implementation.
Finally, a comparison with related work is given,  and we conclude with
a short discussion.

\section{Definitions and Notation}\label{sec:background}

We deal with extended logic programs, which allow for expressing both 
{\em classical negation}%
\footnote{ Note that {\em classical} is a bit of a misnomer since the
operator does not support, for example, contraposition.}
as well as {\em negation as failure\/}
\cite{lifschitz96a}.
We use ``$\neg$'' for classical negation and ``$\nafo$'' for negation as
failure.
Classical negation is also referred to as {\em strong negation}, whilst
negation as failure is termed {\em weak negation}.

Our formal treatment is based on propositional languages.
As usual, a {\em literal}, $L$, is an expression of the form $A$ or $\neg A$,
where $A$ is an atom.
We assume a possibly infinite set of such atoms.
The set of all literals is denoted by $\Lit$.
A literal preceded by the negation as failure sign $\nafo$ is said to be
a {\em weakly negated literal}.
A {\em rule}, $r$, is an ordered pair of the form
\begin{equation}\label{eqn:rule}
L_0\LPif L_1,\dots,L_m,\naf L_{m+1},\dots,\naf L_n,
\end{equation}
where $n\geq m\geq 0$, and each $L_i$ $(0\leq i\leq n)$ is a literal.
The literal $L_0$ is called the {\em head\/} of $r$, and the set
$
\{L_1,\dots,L_m,$ $\naf L_{m+1},\dots,\naf L_n\}
$ 
is the {\em body\/} of $r$.
If $n=m$, then $r$ is a  {\em basic rule\/}; if $n=0$, then $r$ is a {\em fact}.
An (\emph{extended}\/) \emph{logic program}, or simply a \emph{program},
is a finite set of rules.
A program is {\em basic} if all rules in it are basic.
We use $\head{r}$ to denote the head of rule $r$, and $\body{r}$ to
denote the body of~$r$.
Furthermore, let
\(
\pbody{r}
=
\{L_1\commadots L_m\}
\)
and
\(
\nbody{r}
=
\{L_{m+1}\commadots L_n\}
\).
The elements of $\pbody{r}$ are referred to as the {\em prerequisites\/}
of $r$.
We say that a rule $r$ is {\em defeated\/} by a set of literals $X$ iff
$\nbody{r}\cap X\neq\emptyset$.
As well, each literal in $\nbody{r}\cap X$ is said to {\em defeat\/} $r$.

A set of literals $X$ is {\em consistent\/} iff it does not contain a
complementary pair $A$, $\neg A$ of literals.
We say that $X$ is {\em logically closed\/} iff it is either consistent or
equals $\Lit$.
Furthermore, $X$ is {\em closed under\/} a basic program $\Pi$ iff for any
$r\in \Pi$, $\head{r}\in X$ whenever $\body{r}\subseteq X$.
The smallest set of literals which is both logically closed and closed under a
basic program $\Pi$ is denoted by $\Th{\Pi}$.

Let $\Pi$ be a basic program and $X$ a set of literals.
The operator $\To{\Pi}$ is defined as follows:
\[
\T{\Pi}{X} = \{ \head{r}\mid  r\in \Pi, \body{r}\subseteq X\}
\]
if $X$ is consistent, and $\T{\Pi}{X} = \Lit$ otherwise.
Iterated applications of $\To{\Pi}$ are written as $\Tindo{j}{\Pi}$ ($j\geq
0$), where
\(
\Tind{0}{\Pi}{X}=X
\)
and
\(
\Tind{i}{\Pi}{X}=\T{\Pi}{\Tind{i-1}{\Pi}{X}}
\)
for $i\geq 1$.
It is well-known that
\(
\Th{\Pi}=\bigcup_{i\geq 0}\Tind{i}{\Pi}{\emptyset}
\),
for any basic program $\Pi$.

Let $r$ be a rule.
Then $\reductr{r}$ denotes the basic program obtained from $r$ by deleting all
weakly negated literals in the body of $r$, \iec
$\reductr{r}=\head{r}\LPif\pbody{r}$.
The {\em reduct}, $\reduct{\Pi}{X}$, of a program $\Pi$ {\em relative to\/} a
set $X$ of literals is defined by
\[
\reduct{\Pi}{X}
=
\{\reductr{r} \mid \mbox{$r\in \Pi$ and $r$ is not defeated by $X$}\}.
\]
In other words, $\reduct{\Pi}{X}$ is obtained from $\Pi$ by (i) deleting any
$r\in \Pi$ which is defeated by $X$ and (ii) deleting each weakly negated
literal occurring in the bodies of the remaining rules.
We say that a set $X$ of literals is an {\em answer set\/} of a program $\Pi$
iff 
$\Th{\reduct{\Pi}{X}}=X$.
Clearly, for each answer set $X$ of a program $P$, it holds that
$X=\bigcup_{i\geq 0}\Tind{i}{\reduct{\Pi}{X}}{\emptyset}$.
The answer set semantics for extended logic programs has been defined
in~\cite{gellif91a} as a generalization of the stable model
semantics~\cite{gellif88b} for {\em general logic programs\/} (\iec programs
not containing classical negation, $\neg$).
The reduct $\reduct{\Pi}{X}$ is often called the
{\em Gelfond-Lifschitz reduction}. 

The set of all \emph{generating rules} of an answer set $X$ from $\Pi$ is defined as
follows:
\[
\GR{\Pi}{X}
=
\{r\in\Pi\mid\reductr{r}\in\reduct{\Pi}{X}\text{ and }\pbody{r}\subseteq X\}.
\]
That is, each prerequisite of $r$ is in $X$ and $r$ is not defeated by $X$.
Finally, a sequence
\(
\langle r_i\rangle_{i\in I}
\)
of rules is {\em grounded\/} iff, for all $i\in I$,
$\{\head{r_j}\mid j<i\}$ is inconsistent, or else
$\pbody{r_i}\subseteq\{\head{r_j}\mid j<i\}$.

\section{Logic Programs with Preferences}\label{sec:preferences}

\newcommand{\RULE}[4]{#1:&#2&\LPif&#3}

\begin{figure*}
\hrule
\[
\begin{array}{rrclrrcl}
  \RULE{\ap{1}{r}}
       {\head{r}}
       {\applied{\name{r}} }
       {}
  &
  \RULE{\cokt{1}{r}}
       {\ok{\name{r}}}
       {\oko{\name{r}}{\name{r_1}},\ldots,\oko{\name{r}}{\name{r_k}}}
       {}
  \\
  \RULE{\ap{2}{r}}
       {\applied{\name{r}}}
       {\ok{\name{r}},\body{r}}
       {}\hspace{3em}
  &
  \RULE{\cok{2}{r}{r'}}
       {\oko{\name{r}}{\name{r'}}}
       {\naf(\PREC{\name{r}}{\name{r'}})}
       {\text{for }r'\in\Pi}
  \\
  \RULE{\bl{1}{r}{L}}
       {\blocked{\name{r}}}
       {\ok{\name{r}}, \naf L^+}
       {\text{for } L^+\in\pbody{r}}
  &
  \RULE{\cok{3}{r}{r'}}
       {\oko{\name{r}}{\name{r'}}}
       {(\PREC{\name{r}}{\name{r'}}),\applied{\name{r'}}}
       {\text{for }r'\in\Pi}
  \\
  \RULE{\bl{2}{r}{L}}
       {\blocked{\name{r}}}
       {\ok{\name{r}},L^-}
       {\text{for } L^-\in\nbody{r}}
  &
  \RULE{\cok{4}{r}{r'}}
       {\oko{\name{r}}{\name{r'}}}
       {(\PREC{\name{r}}{\name{r'}}),\blocked{\name{r'}}}
       {\text{for }r'\in\Pi}
\end{array}
\]
\[
\begin{array}{rrcl}
  \RULE{t(r,r',r'')}
       {\PREC{\name{r}}{\name{r''}}}
       {\PREC{\name{r}}{\name{r'}},\PREC{\name{r'}}{\name{r''}}}
       {\text{for }r',r''\in\Pi}
  \\
  \RULE{as(r,r')}
       {{\neg(\PREC{\name{r'}}{\name{r}})}}
       {\PREC{\name{r}}{\name{r'}}}
       {\text{for }r'\in \Pi}
\end{array}
\]
\caption{Translated rules $\tau(r)$.}\label{Fig:compile}
\medskip
\hrule
\end{figure*}

A logic program over a propositional language
$\mathcal{L}$ is said to be {\em ordered\/} iff
$\mathcal{L}$ contains the following pairwise disjoint categories:
\begin{itemize}
\item a set $\nameset$ of terms serving as \emph{names} for rules;
  
\item a set $\mathbf{A}$ of regular (propositional) atoms of a program; and
  
\item a set $\mathbf{A}_{\prec}$ of \emph{preference atoms} $\PREC{s}{t}$,
  where $s,t\in\nameset$ are names.
\end{itemize}
For each ordered program $\Pi$, we assume furthermore a bijective%
\footnote{In practice, function $n$ is only required to be injective in order
  to allow for rules not participating in the resultant preference relation.}
function $\namef{\cdot}$ assigning to each rule $r\in\Pi$ a name
$\namef{r}\in\nameset$.
To simplify our notation, we usually write $\name{r}$ instead of
$\namef{r}$
(and we sometimes abbreviate $n_{r_i}$ by $n_i$).
Also, the relation $t=\namef{r}$ is written as $t:r$, leaving the naming
function $\namef{\cdot}$ implicit.
The elements of  ${\mathbf A}_{\prec}$ express preference
relations among rules. 
Intuitively,
$\PREC{\name{r}}{\name{r'}}$ asserts that $r'$ has ``higher'' priority than $r$.
Thus, $r'$ is viewed as having precedence over $r$, \iec
$r'$ should, in some sense, always be considered ``before''
$r$.

Most importantly,
we impose no restrictions on the occurrences of preference atoms.
This allows for expressing preferences in a very flexible, dynamic way.
For instance, we may specify
\[
\name{r}\prec\name{r'}\LPif p, \naf q
\]
where $p$ and $q$ may themselves be (or rely on) preference atoms.

A special case is given by programs containing preference atoms only among
their facts.
We say that a logic program $\Pi$ over $\mathcal{L}$ is
\emph{statically ordered}
if it is of the form
\(
\Pi=\Pi'\cup\Pi''
\),
where
$\Pi'$ is an ordered
logic program over $\mathcal{L}\setminus{\mathbf A}_{\prec}$
and
\(
\Pi''
\subseteq
\{(\PREC{\name{r}}{\name{r'}})\LPif{}\mid r,r'\in\Pi'\}.
\)
The static case can be regarded as being induced from an external order
``$<$'', where the relation \PRECM{r}{r'} between two rules holds iff
the fact $(\PREC{\name{r}}{\name{r'}})\LPif$ is included in the ordered
program.
We make this explicit by denoting a statically ordered program $\Pi$ as a pair
\(
(\Pi',<)
\),
representing the program
\(
\Pi'\cup\{(\PREC{\name{r}}{\name{r'}})\LPif{}\mid\PRECM{r}{r'}\}
\).
This static concept of preference corresponds in fact to most previous
approaches to preference handling in logic programming and nonmonotonic
reasoning,
where the preference information is specified as a fixed relation at the
meta-level (see, \egc \cite{baahol93a,brewka94a,zhafoo97a,breeit99a}).

Our approach provides a mapping $\mathcal{T}$ that transforms an ordered
logic program $\Pi$ into a regular logic program $\mathcal{T}(\Pi)$, such that
the preferred answer sets of $\Pi$ are given by the (regular) answer sets of
$\mathcal{T}(\Pi)$.
Intuitively, the translated program $\mathcal{T}(\Pi)$ is constructed in such
a way that the ensuing answer sets respect the inherent preference information
induced by the given program $\Pi$ (see Theorems~\ref{thm:priority:implementing}
and~\ref{thm:priority:preserving} below).
This is achieved by adding sufficient control elements to the
rules of $\Pi$ which guarantee that successive rule applications are in
accord with the intended order.

Given the relation $\PREC{\name{r}}{\name{r'}}$, we want to ensure that $r'$ is
considered before $r$, in the sense that, for a given answer set $X$, rule
$r'$ is known to be applied or defeated \emph{ahead of} $r$ with respect to the
grounded enumeration of generating rules of $X$.
We do this by first translating rules so that the order of rule application
can be explicitly controlled.
For this purpose, we need to be able to detect when a rule has been applied
or when a rule is defeated;
as well we need to be able to control the application of a rule based on
other antecedent conditions.
For a rule $r$, there are two cases for it not to be applied:
it may be that some literal in $\pbody{r}$ does not appear in the answer set,
or it may be that the negation of a literal in $\nbody{r}$ is in the answer
set.
For detecting this case, we introduce,
for each rule $r$ in the given program $\Pi$, a new, special-purpose atom
$\blocked{\name{r}}$.
Similarly, we introduce a special-purpose atom $\applied{\name{r}}$
to detect the case where a rule has been applied.
For controlling application of rule $r$ we introduce the 
atom $\ok{\name{r}}$.
Informally, we conclude that it is $\oksym$ to apply a rule just if it is
$\oksym$ with respect to every $\prec$-greater rule;
for such a $\prec$-greater rule $r'$, this will be the case just when $r'$ is
known to be blocked or applied.

More formally, given an ordered program $\Pi$ over $\mathcal{L}$,
let $\mathcal{L}^+$ be the language obtained from $\mathcal{L}$ by adding,
for each $r,r'\in\Pi$, new pairwise distinct propositional atoms
$\applied{\name{r}}$, $\blocked{\name{r}}$, $\ok{\name{r}}$, and
$\oko{\name{r}}{\name{r'}}$.
Then, our translation $\mathcal{T}$ maps an ordered program $\Pi$ over
$\mathcal{L}$ into a regular program $\mathcal{T}(\Pi)$ over $\mathcal{L}^+$
in the following way.
%
\begin{definition}\label{def:compilation}
Let $\Pi = \{r_1\commadots r_k\}$ be an ordered logic program over
$\mathcal{L}$. For each $r\in\Pi$\/, let $\tau(r)$ be the collection of 
rules depicted in Figure~\ref{Fig:compile}, where
$L^+\in\pbody{r}$,
$L^-\in\nbody{r}$,
and
$r',r'' \in \Pi$. 
Then,
the logic program $\mathcal{T}(\Pi)$ over $\mathcal{L}^+$ is given by $\bigcup_{r\in\Pi}\tau(r)$.

\end{definition}
%
The first four rules of Figure~\ref{Fig:compile} express applicability 
and blocking conditions of the original rules:
For each rule $r\in\Pi$, we obtain two rules, \ap{1}{r} and \ap{2}{r}, along
with $n$ rules of the form \bl{1}{r}{L}, \bl{2}{r}{L},
where $n$ is the number of
literals in the body.
The second group of rules encodes the strategy for handling preferences.
The first of these rules, \cokt{1}{r}, ``quantifies'' over the rules in $\Pi$.
This is necessary when dealing with dynamic preferences since preferences may
vary depending on the corresponding answer set.
The three rules \cok{2}{r}{r'}, \cok{3}{r}{r'}, and \cok{4}{r}{r'} specify
the pairwise dependency of rules in view of the given preference ordering:
For any pair of rules $r$, $r'$ with $\name{r}\prec\name{r'}$,
we derive $\oko{\name{r}}{\name{r'}}$ whenever $\name{r}\prec\name{r'}$ fails 
to hold, or whenever either $\applied{\name{r'}}$ or
$\blocked{\name{r'}}$ is true.
This allows us to derive $\ok{\name{r}}$, indicating that $r$ may potentially
be applied whenever we have for all $r'$ with $\name{r}\prec\name{r'}$ that
$r'$ has been applied or cannot be applied.
It is important to note that this is only one of many strategies for dealing
with preferences: different strategies are obtainable by changing the
specification of \oko{\cdot}{\cdot}.
Finally, we note that our implementation represents the second group of rules
in terms of  {\em four rule schemas\/} (using variables), where the first 
one depends on
the number of {\em names\/}
(as opposed to the number of rules; cf.\ Definition~\ref{def:compilation}).

We have the following characterisation of \emph{preferred answer sets}.
\begin{definition}
  Let $\Pi$ be an ordered logic program over language $\mathcal{L}$ 
  and $X$ a set of literals.
  We say that $X$ is a preferred answer set of\/ $\Pi$
  iff
  $X=Y\cap\mathcal{L}$ for some answer set $Y$ of\/ $\mathcal{T}(\Pi)$.
\end{definition}

In what follows, answer sets of standard (\iec unordered) logic programs are
also referred to as \emph{regular answer sets}.


As an illustration of our approach, consider the following program $\Pi$:
\[
\begin{array}{rcrcl}
  r_1 & = & \neg a             &\LPif&
  \\
  r_2 & = & \phantom{\neg} b   &\LPif&\neg a, \naf c
  \\
  r_3 & = & \phantom{\neg} c   &\LPif& \naf b
  \\
  r_4 & = & n_3\prec n_2       &\LPif& \naf d
\end{array}
\]
where $\name{i}$ denotes the name of rule $r_i$ $(i=1\commadots 4)$.
This program has two regular answer sets,
one containing
\(
b
\)
and the other containing
\(
c
\);
both contain $\neg a$ and $n_3\prec n_2$.
However, only the first is a preferred answer set.
To see this, observe that for any
\(
X\subseteq\{\head{r}\mid r\in\mathcal{T}(\Pi)\}
\),
we have
\(
n_i\prec n_j\not\in X
\)
for each $(i,j)\neq (3,2)$.
We thus get for such $X$ and $i,j$ that
\(
\oko{\name{i}}{\name{j}}\in\Tind{1}{\reduct{\mathcal{T}(\Pi)}{X}}{\emptyset}
\)
by (reduced) rules $\cok{2}{r_i}{r_j}^+$, and so
\(
\ok{\name{i}}\in\Tind{2}{\reduct{\mathcal{T}(\Pi)}{X}}{\emptyset}
\)
via rule $\cokt{1}{r_i}^+=\cokt{1}{r_i}$.
Analogously, we get
\(
\applied{\name{1}},\applied{\name{4}},
\neg a, n_3\prec n_2
\).
Now consider the following rules from $\mathcal{T}(\Pi)$:
\[
\begin{array}{rcrcl}
  \ap{2}{r_2}&:&  \applied{\name{2}}&\LPif& \ok{\name{2}},\neg a, \naf c 
  \\
  \bl{1}{r_2}{\neg a}&:&  \blocked{\name{2}}&\LPif& \ok{\name{2}}, \naf \neg a
  \\
  \bl{2}{r_2}{c}&:&  \blocked{\name{2}}&\LPif& \ok{\name{2}},c 
  \\[1ex]
  \ap{2}{r_3}&:&  \applied{\name{3}}&\LPif& \ok{\name{3}}, \naf b 
  \\
  \bl{2}{r_3}{c}&:&  \blocked{\name{3}}&\LPif& \ok{\name{3}},b 
  \\[1ex]
  \cok{3}{r_3}{r_2}&:&  \oko{\name{3}}{\name{2}}&\LPif& (\PREC{\name{3}}{\name{2}}), \applied{\name{2}}
  \\
  \cok{4}{r_3}{r_2}&:&  \oko{\name{3}}{\name{2}}&\LPif& (\PREC{\name{3}}{\name{2}}), \blocked{\name{2}}
\end{array}
\]
Given $\ok{\name{2}}$ and $\neg a$, rule \ap{2}{r_2} leaves us with the
choice between
\(
c\not\in X
\)
or
\(
c\in X
\).
First, assume $c\not\in X$.
We get \applied{\name{2}} from
\(
\ap{2}{r_2}^+\in\mathcal{T}(\Pi)^X
\).
Hence, we get $b$, \oko{\name{3}}{\name{2}}, and finally \ok{\name{3}},
which results in \blocked{\name{3}} via \bl{2}{r_3}{c}.
Omitting further details, this yields an answer set containing $b$ while 
excluding $c$.
Second, assume $c\in X$.
This eliminates \ap{2}{r_2} when turning $\mathcal{T}(\Pi)$ into $\mathcal{T}(\Pi)^X$.
Also, \bl{1}{r_2}{\neg a} is defeated since $\neg a$ is derivable.
\bl{2}{r_2}{c} is inapplicable, since $c$ is only derivable
(from \applied{\name{3}} via \ap{1}{r_3})
in the presence of \ok{\name{3}}.
But \ok{\name{3}} is not derivable since neither
\applied{\name{2}} nor \blocked{\name{2}} is derivable.
Since this circular situation is unresolvable, there is no preferred answer
set containing~$c$.

\section{Properties of the Approach}\label{sec:properties}

Our first result ensures that the dynamically generated preference information
enjoys the usual properties of strict orderings.
To this end, we define the following relation: 
for each set $X$ of literals and every $r,r'\in\Pi$, the relation
$\PRECi{X}{r}{r'}$ holds iff $\PREC{\name{r}}{\name{r'}}\in X$.
%
\begin{theorem}\label{thm:strict:order}
  Let $\Pi$ be an ordered logic program and $X$ a consistent answer set
  of~$\mathcal{T}(\Pi)$.
  Then $\PRECio{X}$ is a strict partial order.
  Moreover, if\/ $\Pi$ has only static preferences,
  then $\PRECio{X}=\PRECio{Y}$, for any answer set $Y$ of $\mathcal{T}(\Pi)$.
\end{theorem}

The following results shed light on the functioning induced by translation
$\mathcal{T}$; they elaborate upon the logic programming operator
\To{\mathcal{T}(\Pi)}~:
%
\begin{theorem}\label{thm:rigid:results}
  Let $X$ be a consistent answer set of
  \(
  \mathcal{T}(\Pi)
  \) 
  for an ordered program $\Pi$, 
  and let $\Omega=
  \reduct{\mathcal{T}(\Pi)}{X}
  $. 
  Then,
  for any 
  \(
  r\in \Pi
  \):
  \begin{enumerate}
  \item  \label{l:rigid:results:2}
    $\ok{\name{r}}\in X$;
  \item \label{l:rigid:results:3}
    $\applied{\name{r}}\in X$
    iff\/
    $\blocked{\name{r}}\not\in X$;
  \item   \label{l:rigid:results:4}
    if $r$ is not defeated by $X$, $\ok{\name{r}}\in \Tind{i}{\Omega}{\emptyset}$, 
    and
    $\pbody{r}\subseteq\Tind{j}{\Omega}{\emptyset}$, then 
    $\applied{\name{r}}\in\Tind{\max(i,j)+1}{\Omega}{\emptyset}$;
  \item   \label{l:rigid:results:5}
    $\ok{\name{r}}\in \Tind{i}{\Omega}{\emptyset}$
    and
    $\pbody{r}\not\subseteq X$
    implies 
    $\blocked{\name{r}}\in \Tind{i+1}{\Omega}{\emptyset}$;
  \item   \label{l:rigid:results:6}
    if $r$ is defeated by $X$ and
    $\ok{\name{r}}\in \Tind{i}{\Omega}{\emptyset}$, then
    $\blocked{\name{r}}\in \Tind{j}{\Omega}{\emptyset}$
    for some $j>i$;
  \item   \label{l:rigid:results:7}
    $\ok{\name{r}}\not\in \Tind{i}{\Omega}{\emptyset}$
    implies 
    $\applied{\name{r}}\not\in \Tind{j}{\Omega}{\emptyset}$
    and
    $\blocked{\name{r}}\not\in \Tind{k}{\Omega}{\emptyset}$
    for all $j,k< i+2$.
  \end{enumerate}
\end{theorem}

The next result shows that the translated rules are considered in accord to
the partial order induced by the given preference relation:
%
\begin{theorem}\label{thm:priority:implementing}
  Let $\Pi$ be an ordered logic program, 
  $X$ a consistent answer set of $\mathcal{T}(\Pi)$,
  and
  \(
  \langle r_i\rangle_{i\in I}
  \)
  a grounded enumeration
  of the set \GR{\mathcal{T}(\Pi)}{X} of generating rules of $X$ 
  from ${\cal T}(\Pi)$.
  Then, for all
  \(
  r,r'\in \Pi
  \):

  \[
  \text{If } \PRECi{X}{r}{r'}, \text{ then } j<i,
  \]
  for all $r_i$
  equaling 
  $\ap{k}{r}$
  or $\bl{k}{r}{L}$, and
  some
  $r_j$
  equaling 
  $\ap{k'}{r'}$
  or
  $\bl{k'}{r'}{L'}$,
  with
  $k,k'=1,2$, 
  $L\in\body{r}$,
  and
  $L'\in\body{r'}$.
\end{theorem}

For static preferences, our translation $\mathcal{T}$ amounts to
selecting the answer sets of the underlying unordered program that comply
with the ordering, $<$.
%
\begin{definition}\label{def:priority:preserving}
  Let $(\Pi,<)$ be a statically ordered program.
  An answer set $X$ of\/ $\Pi$ is called  $<$-preserving if
  $X$ is either inconsistent, or else 
  there exists a grounded enumeration
  \(
  \langle r_i\rangle_{i\in I}
  \)
  of $\GR{\Pi}{X}$
  such that, for every $i,j\in I$, we have that:
  \begin{enumerate}
    \item if $r_i<r_j$, then $j<i$;
          \ and
    \item if
          $\PRECM{r_i}{r'}$
          and 
          \(
          r'\in {\Pi\setminus\GR{\Pi}{X}},
          \)
          then
          $\pbody{r'}\not\subseteq X$
          or
          $r'$ is defeated by 
          the set $\{\head{r_j}\mid j<i\}$.
  \end{enumerate}
\end{definition}
%
The next result furnishes semantical underpinnings for statically ordered
programs;
it provides a correspondence between preferred answer sets and regular answer
sets of the original program:
%
\begin{theorem}\label{thm:priority:preserving}
  Let $(\Pi,<)$ be a statically ordered logic program
  and
  $X$ a set of literals.
  Then,
  $X$ is a preferred answer set of
  \(
  (\Pi,<)
  \)
  iff
  $X$ is a $<$-preserving answer set of\/ $\Pi$.
\end{theorem}
%
This gives rise to the following corollary:
%
\begin{corollary}\label{thm:rigid:standard}
  Let $(\Pi, <)$ and $X$ be as 
  in Theorem~\ref{thm:priority:preserving}.
  If
  $X$ is a preferred answer set of\/
  \(
  (\Pi, <)
  \),
  then
  $X$ is an answer set of\/
  \(
  \Pi
  \).
\end{corollary}
%
Note that the last two results
have no counterparts in the general (dynamic) case, due to the lack of a
regular answer set of the original program.
The preference information is only fully available in the answer sets of the
translated program (hence the restriction of the notion of $<$-preservation to
the static case).

Also, if no preference information is present, our approach is equivalent to
standard answer set semantics.
Moreover, the notions of statically ordered and (dynamically) ordered programs
coincide in this case.
%
\begin{theorem}\label{thm:rigid:reiter:all}
  Let $\Pi$ be a logic program over ${\cal L}$ and $X$ a set of literals. 
  If\/ $\Pi$ contains no preference information,
  \ie if ${\cal L}\cap{\mathbf A}_\prec=\emptyset$, 
  then
  the following statements are equivalent:
  \begin{enumerate}
  \item $X$ is a preferred answer set of
    statically ordered logic program $(\Pi,\emptyset)$;
  \item $X$ is a preferred answer set of ordered logic program $\Pi$; 
  \item $X$ is a regular answer set of logic program $\Pi$.
\end{enumerate}
\end{theorem}

Recently, Brewka and Eiter~\cite{breeit99a} suggested two principles,
simply termed {\em Principle~I\/} and {\em Principle~II\/}, which,
they argue, any defeasible rule system handling preferences should satisfy.
The next result shows that our approach obeys these  principles. 
However, since the original formulation of Principle~I and II is
rather generic---motivated by the aim to cover as many different 
approaches as possible---we must instantiate them 
in terms of our formalism.
It turns out that Principle~I is only suitable for statically ordered
programs, whilst Principle~II admits two guises, one for statically
ordered programs, and another one for (dynamically) ordered programs.

Principles I and II, formulated for our approach, are as follows:

\begin{description}
  \item[Principle I.] 
    Let $(\Pi,<)$ be a statically ordered logic program, 
    and let $X_1$ and $X_2$ be two (regular) answer sets of $\Pi$ generated
    by $R\cup\{r_1\}$ and $R\cup\{r_2\}$, respectively, where $r_1,r_2\not\in R$.
    If $\PRECM{r_1}{r_2}$, then $X_1$ is not a
    preferred answer set of $(\Pi,<)$.

  \item[Principle II-S (Static Case).] 
    Let $X$ be a preferred answer set of
    statically ordered logic program $(\Pi,<)$, let $r$ be a rule wherein
    $\pbody{r}\not\subseteq X$, and let  $<'$ 
    be a strict partial order which agrees with $<$ on rules from $\Pi$.
    Then, 
    $X\cup A$ 
    is an answer set of 
    $(\Pi\cup\{r\},<')$,
    where 
    \[
    A=\{\PREC{\name{r}}{\name{s}}\mid 
         r<'s\}\cup\{\neg(\PREC{\name{s}}{\name{r}})\mid 
         r<'s\}.\footnotemark
    \]
    \footnotetext{The inclusion of 
      $A$ 
      is necessary because we encode the preference 
      information at the object level.}
  \item[Principle II-D (Dynamic Case).] 
    Let $X$ be a preferred answer set of a
    (dynamically) ordered logic program $\Pi$, and let $r$ be a rule such that 
    $\pbody{r}\not\subseteq X$.
    Then, 
    $X$ 
    is an answer set of 
    $\Pi\cup\{r\}$.

\end{description}
%
\begin{theorem}
Statically ordered logic programs obey Principles I and II-S. 
Furthermore, ordered logic programs enjoy Principle II-D.
\end{theorem}

Observe that, since transformation $\cal T$ is clearly polynomial 
in the size of ordered logic programs,  
and because of Theorem~\ref{thm:rigid:reiter:all}, the 
complexity of our approach is inherited from the complexity of standard answer set
semantics in a straightforward way.
We just note the following result:
%
\begin{theorem}
Given an ordered program $\Pi$, checking whether $\Pi$ has a preferred answer
set is NP-complete.
\end{theorem}

\section{Further Issues and Refinements}\label{sec:extensions}

In this section, we sketch the range of applicability and point out
distinguishing features of our approach.
We briefly mention two points concerning expressiveness, and then 
sketch how we can deal with preferences over sets of rules.
Lastly, we refer to the implementation of our approach.

First, we draw the reader's attention to the expressive power offered by
dynamic preferences in connection with variables in the input language,
such as
\begin{equation}
\label{eqn:genpref}
n_1(x)\prec n_2(y)\LPif p(y), \naf (x=c),
\end{equation}
where $n_1(x),n_2(y)$ are names of rules containing the variables $x$ and $y$,
respectively.
Although such a rule represents only its set of ground instances,
it is actually a much more concise specification.
Also, since most other approaches employ static preferences of the form
\(
n_1(x)\prec n_2(y)\LPif,
\)
such approaches would necessarily have to express (\ref{eqn:genpref})
as an enumeration of static ground preferences rather than a single rule.

Second, we note that transformation $\mathcal{T}$ is also applicable to
disjunctive logic programs (where rule heads are disjunctions of literals).
To see this, observe that the transformed rules unfold the conditions
expressed in the body of the rules, while the rules' head remain untouched,
as manifested by rule
\(
\ap{1}{r} 
\).

Third, we have extended the approach to allow for preferences between
sets of rules.
Although we do not include
a full discussion here, we remark that this extension has also
been implemented (see below).
In order to refer to sets of rules, the language is adjoined by a set $\mameset$
of terms serving as names for sets of rules, and, in addition,
the set ${\mathbf A}_{\prec}$ may now include atoms of
the form $\PREC{m}{m'}$ with $m,m'\in\mameset$.
Accordingly, {\em set-ordered programs\/} contain preference information
between names of sets.
Informally, set $M$ of rules is applicable iff all its members are applicable.
Consequently, if $M'$ is preferred over $M$, then $M$ is considered after 
{\em all\/} rules in $M'$ are found to be applicable, or some rule in $M'$ 
is found to be inapplicable. 
As before, set-ordered programs are translated into standard logic programs,
where suitable control elements $\ok{\cdot}$, $\blocked{\cdot}$, and
$\applied{\cdot}$, ranging over names of sets, take care of the intended 
ordering information.

As an example, consider where in buying a car one ranks the price ($e$)
over safety features ($s$) over power ($p$),
but safety features together with power is ranked over price.
Taking
\(
r_x = x\LPif\naf\neg x
\)
for $x\in\{e,s,p\}$,
we can write this (informally) as:
\[
m_1    : \left\{ r_p \right\}
\; < \;
m_2    : \left\{ r_s \right\}
\; < \;
m_3    : \left\{ r_e \right\}
\; < \;
m_4: \left\{ r_p, r_s \right\}
\]
The terms $m_1$, $m_2$, $m_3$, and $m_4$ are names of sets of rules.
If we were given only that not all desiderata can be satisfied
(\ie $\LPif p, e, s$)
then we could apply the rules in the set (named) $m_4$ and conclude
that $p$ and $s$ can be met.
Furthermore, sets of rules are described extensionally by means of atoms $\inp{\cdot}{\cdot}$.
Thus, the set $m_4:\{r_p,r_s\}$ is captured by
\(
\inp{n_p}{m_4}\LPif
\)
and
\(
\inp{n_s}{m_4}\LPif
\).
Accordingly, we have $\inp{n_p}{m_1}\LPif$, $\inp{n_s}{m_2}\LPif$, and
$\inp{n_e}{m_3}\LPif$.
Given rules $r_e,r_p,r_s$ and the previous facts about $\insym$,
the specification of our example is completed by the preferences
$m_i\prec m_{i+1}\LPif$ for $i=1,2,3$.

Lastly, the approach has been implemented in Prolog and serves as a
front-end to the logic programming systems \texttt{dlv} and \texttt{smodels}.
The current prototype is available at
\begin{quote}
\texttt{http://www.cs.uni-potsdam.de/\~{}torsten/plp/}.
\end{quote}
This URL contains also diverse examples taken from the literature.
Both the dynamic approach to (single) preferences and the set-based approach
have been implemented. 
We note also that the implementation differs from the approach described
here in two respects: first, the translation applies to named rules only, \iec it
leaves unnamed rules unaffected; and second, it  provides a module 
which admits the specification of rules containing variables, whereby rules of
this form are processed by applying an additional grounding step.
A more detailed account regarding the implemented front-end
can be found in \cite{delschtom:sys00}.

\section{Comparison with Related Work}\label{sec:discussion}

Dealing with preferences on rules seems to necessitate a two-level approach.
This in fact is a characteristic of many approaches found in the literature.
The majority of these approaches treat preference at the meta-level by
defining alternative semantics.
\cite{brewka96a} proposes a modification of well-founded semantics in which
dynamic preferences may be given for rules employing $not$.
\cite{zhafoo97a} and
\cite{breeit99a} propose different prioritized versions of answer set
semantics.
In \cite{zhafoo97a} static preferences are addressed first, by defining the
{\em reduct} of a logic program $\Pi$, which is a subset of $\Pi$ that is
most preferred.
For the following example, their approach gives two answer sets (one with $p$
and one with $\neg p$) which seems to be counter-intuitive; ours in contrast 
has a single answer set containing
$\neg p$.
\begin{eqnarray*}
r_1
&:&
p \LPif{} \naf q_1
\\
r_2
&:& 
\neg p \LPif{} \naf q_2
\\
&&
r_1 < r_2
\end{eqnarray*}
Moreover, the dynamic case is addressed by specifying a transformation of a
dynamic program to a set of static programs.

Brewka and Eiter \cite{breeit99a} address static preferences on rules in
extended logic programs.
They begin with a strict partial order on a set of rules, but
define preference with respect to total orders that conform to the original
partial order.
Preferred answer sets are then selected from among the collection of answer sets
of the (unprioritised) program.
In contrast, we deal only with the original partial order, which is
translated into the object theory.
As well, only preferred extensions are produced in our approach;
there is no need for meta-level filtering of extensions.

A two-level approach is also found in \cite{gelson97a}, where a
methodology for directly encoding preferences in logic programs is proposed.
The ``second-order flavour'' of this approach stems from the reification of
rules and preferences.
For example, a rule
\(
p\LPif{} r, \neg s, \naf q
\)
is expressed by the formula
\(
\mathit{default}(n, p, [r, \neg s], [q])
\)
where $n$ is the name of the rule.
The Prolog-like list notation $[r, \neg s]$ and $[q]$ raises the possibility
of an infinite Herbrand universe;
this is problematic for systems like
\texttt{smodels} and \texttt{dlv} that rely on finite Herbrand universes.

\section{Conclusion}\label{sec:conclusion}

We have described an approach for compiling preferences into logic programs
under the answer set semantics.
An ordered logic program, in which preferences appear in the program rules,
is transformed into a second, extended logic program wherein the preferences
are respected, in that the answer sets obtained in the transformed theory
correspond with the preferred answer sets of the original theory.
In a certain sense, our transformation can be regarded as an axiomatisation
of (our interpretation of) preference.
Arguably then, we describe a general {\em methodology} for uniformly
incorporating preference information in a logic program.
In this approach, we avoid the two-level structure of previous work.
While the previous ``meta-level'' approaches must commit themselves to a
semantics and a fixed strategy, our approach (as well as that of
\cite{gelson97a}) is very flexible with respect to changing strategies, and
is open for adaptation to different semantics and different concepts of
preference handling.

The approach is easily restricted to reflect a {\em static} ordering
in which preferences are external to a logic program.
We also indicated how the approach can be extended to deal with preferences
among sets of rules.
Finally, this paper demonstrates that our approach is easily implementable;
indeed, we have developed a compiler, as a front-end for
\texttt{dlv} and \texttt{smodels}.


\end{document}